\documentclass[10pt,twocolumn,letterpaper]{article}

\usepackage{cvpr}
\usepackage{times}
\usepackage{epsfig}
\usepackage{graphicx}
\usepackage{amsmath}
\usepackage{amssymb}
\usepackage{array}
\usepackage{pdfpages}
\usepackage{graphbox}
\usepackage{algorithm}
\usepackage{algpseudocode}
\usepackage{tabu,multirow}
\usepackage{booktabs}
\usepackage[percent]{overpic}

\usepackage[pagebackref=true,breaklinks=true,letterpaper=true,colorlinks,bookmarks=false]{hyperref}

\def\eg{\emph{e.g.}}

\def\ie{\emph{i.e.}}
\def\etal{\emph{et al}}

\def\netname{\emph{MADNet}}
\def\extendednetname{\underline{M}odularly \underline{AD}aptive \underline{Net}work}
\def\algoname{\emph{MAD}}
\def\extendedalgoname{\underline{M}odular \underline{AD}aptation}

\def\kitti{KITTI}

\cvprfinalcopy 

\def\cvprPaperID{2901} 

\ifcvprfinal\fi
\begin{document}

\title{Real-time self-adaptive deep stereo}

\author{Alessio Tonioni, Fabio Tosi, Matteo Poggi, Stefano Mattoccia, Luigi di Stefano\\
Department of Computer Science and Engineering (DISI)\\
University of Bologna, Italy\\
{\tt\small \{alessio.tonioni, fabio.tosi5, m.poggi, stefano.mattoccia, luigi.distefano \}@unibo.it}
}

\maketitle
\ifcvprfinal\thispagestyle{empty}\fi

\begin{abstract}

Deep convolutional neural networks trained end-to-end are the state-of-the-art methods to regress dense disparity maps from stereo pairs. These models, however, suffer from a notable decrease in accuracy when exposed to scenarios significantly different from the training set (\eg{,} real vs synthetic images, etc.). We argue that it is extremely unlikely to gather enough samples to achieve effective training/tuning in any target domain, thus making this setup impractical for many applications. Instead, we propose to perform unsupervised and continuous online adaptation of a deep stereo network, which allows for preserving its accuracy in any environment. However, this strategy is extremely computationally demanding and thus prevents real-time inference. We address this issue introducing a new lightweight, yet effective, deep stereo architecture, \extendednetname{} (\netname{}), and developing a \extendedalgoname{} (\algoname{}) algorithm, which independently trains sub-portions of the network. By deploying \netname{} together with \algoname{} we introduce the first real-time self-adaptive deep stereo system enabling competitive performance on heterogeneous datasets. Our code is 
publicly available at \url{https://github.com/CVLAB-Unibo/Real-time-self-adaptive-deep-stereo}.
\end{abstract}

\section{Introduction}
\label{sec:intro}

Many key tasks in computer vision rely on the availability of dense and reliable 3D reconstructions of the sensed environment. Due to high precision, low latency and affordable costs, passive stereo has proven particularly amenable to depth estimation in both indoor and outdoor set-ups. Following the groundbreaking work by Mayer \etal{} \cite{mayer2016large}, current state-of-the-art stereo methods rely on deep convolutional neural networks (CNNs) that take as input a pair of left-right frames and directly regress a dense disparity map. In challenging real-world scenarios, like the popular \kitti{} benchmarks \cite{KITTI_2012,KITTI_2015}, these networks turn out to be more effective, and sometimes faster, than \emph{traditional} algorithms. 

As recently highlighted in \cite{Tonioni_2017_ICCV,pang2018zoom}, learnable models suffer from loss in performance when tested on unseen scenarios due to the domain shift between training and testing data - often synthetic and real, respectively. Good performance can be regained by fine-tuning on \textit{few} annotated samples from the target domain. Yet, obtaining groundtruth labels requires the use of costly active sensors (\eg, LIDAR) and  noise removal by expensive manual intervention or post-processing \cite{Uhrig2017THREEDV}. Recent works \cite{Tonioni_2017_ICCV,pang2018zoom,zhou2017unsupervisedStereo,godard2017unsupervised,zhang2018activestereonet} proposed to overcome the need for labels with unsupervised losses that require only stereo pairs from the target domain. Although effective, these techniques are inherently limited by the number of samples available at training time. Unfortunately, for many tasks, like autonomous driving, it is unfeasible to acquire, in advance, samples from all possible deployment domains (\eg, every possible road and/or weather condition).

\begin{figure*}
	\setlength{\tabcolsep}{1pt}
	\center
	\begin{tabular}{cccc}
		\vspace{1pt}
		(a)&
		\includegraphics[align=c,width=0.30\textwidth]{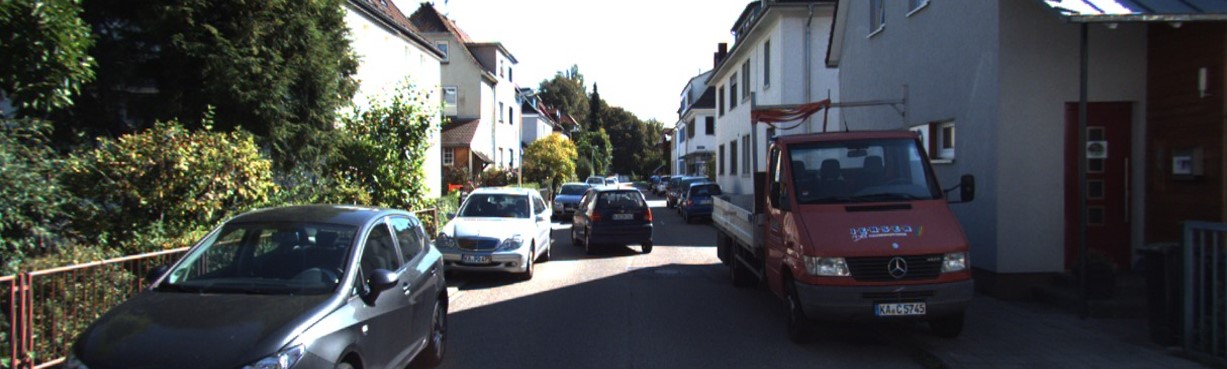} & 
		\includegraphics[align=c,width=0.30\textwidth]{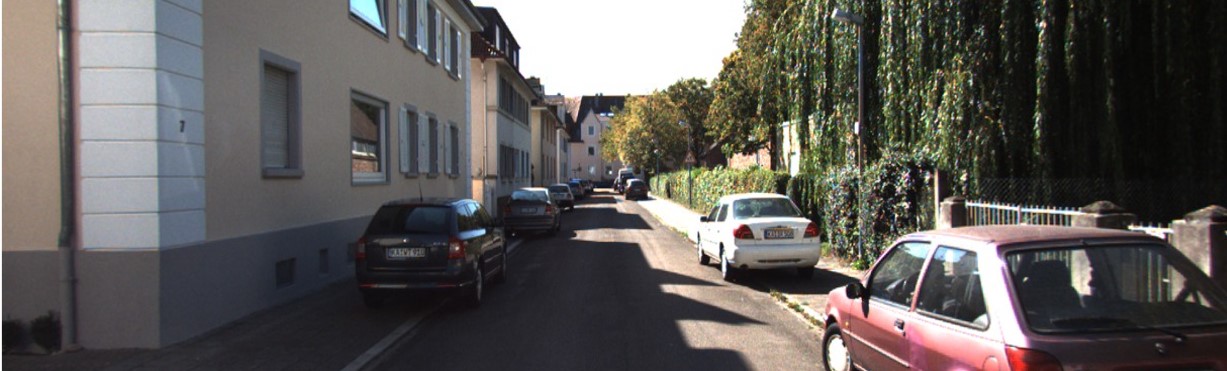} & 
		\includegraphics[align=c,width=0.30\textwidth]{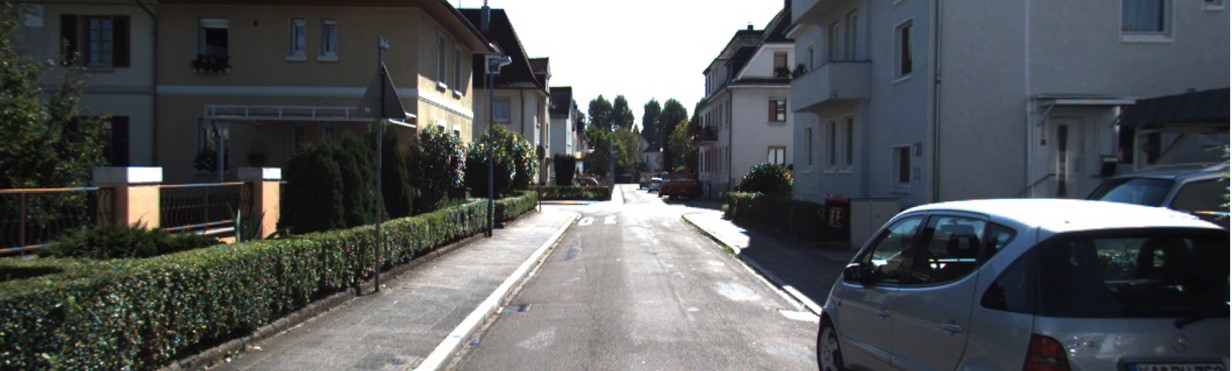} \\
		
		\vspace{1pt}
		(b)&
		\includegraphics[align=c,width=0.30\textwidth]{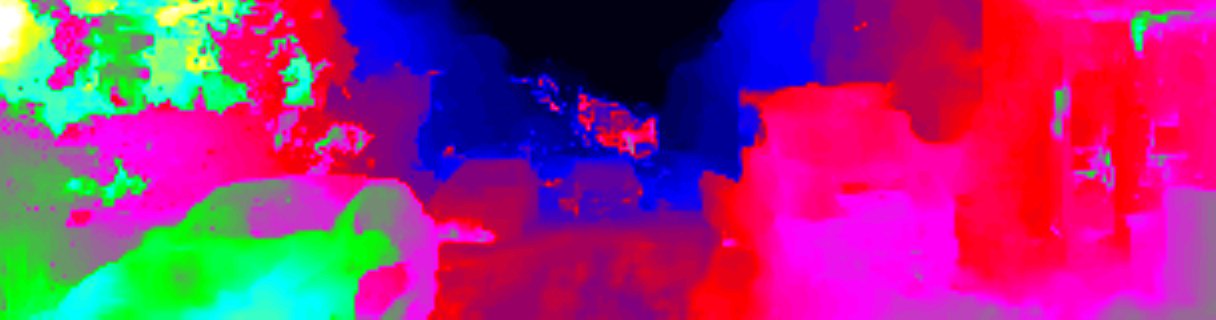} &
		\includegraphics[align=c,width=0.30\textwidth]{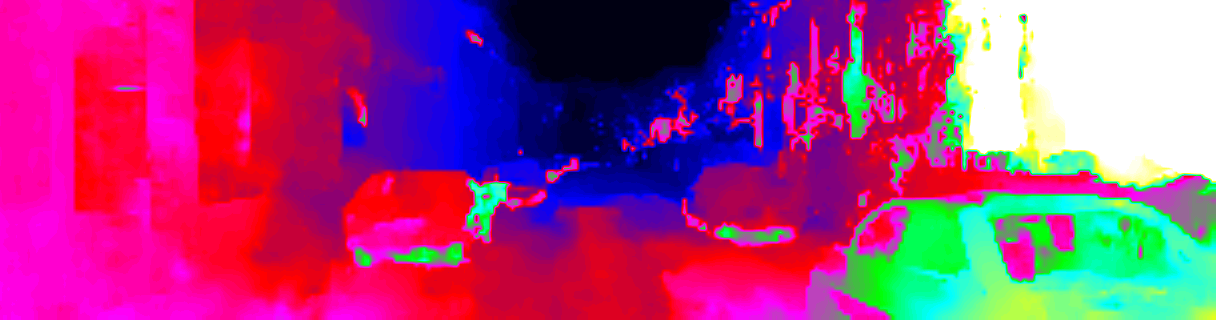} &
		\includegraphics[align=c,width=0.30\textwidth]{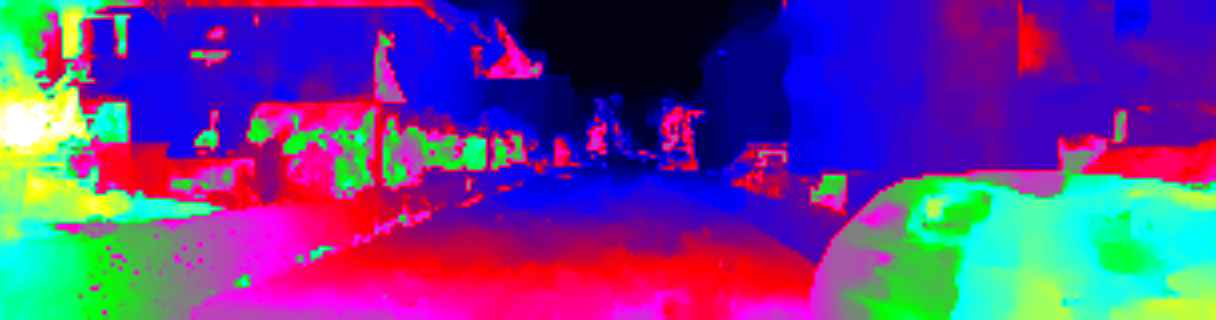}\\
		
		\vspace{1pt}
		(c)&
		\includegraphics[align=c,width=0.30\textwidth]{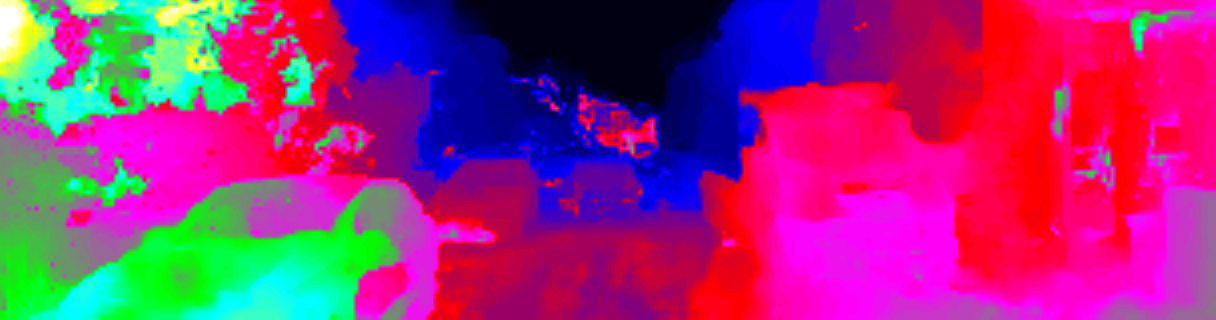} &
		\includegraphics[align=c,width=0.30\textwidth]{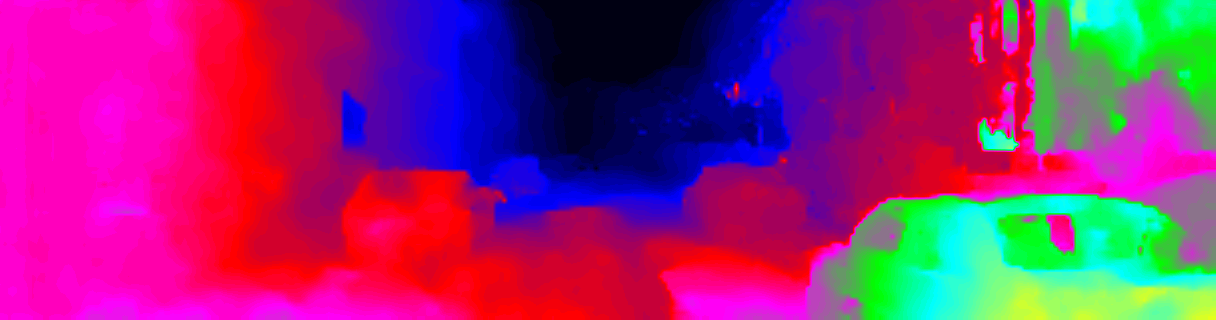} &
		\includegraphics[align=c,width=0.30\textwidth]{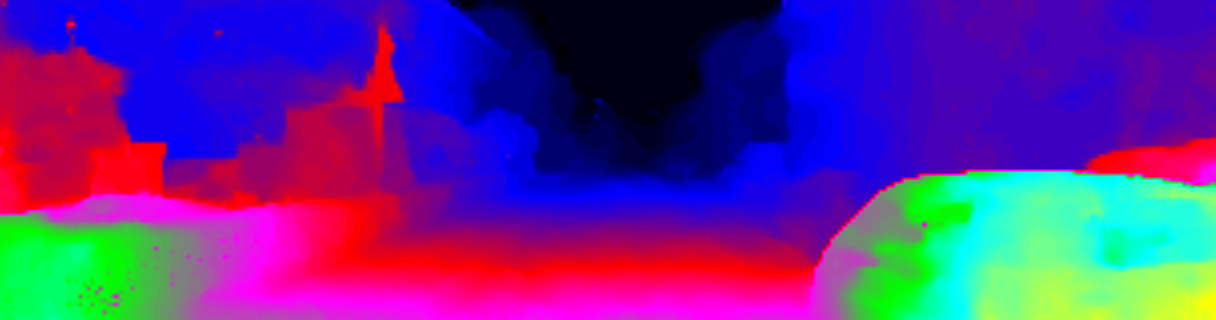}\\
		
		\vspace{1pt}
		(d)&
		\includegraphics[align=c,width=0.30\textwidth]{C_disparity_0.png} &
		\includegraphics[align=c,width=0.30\textwidth]{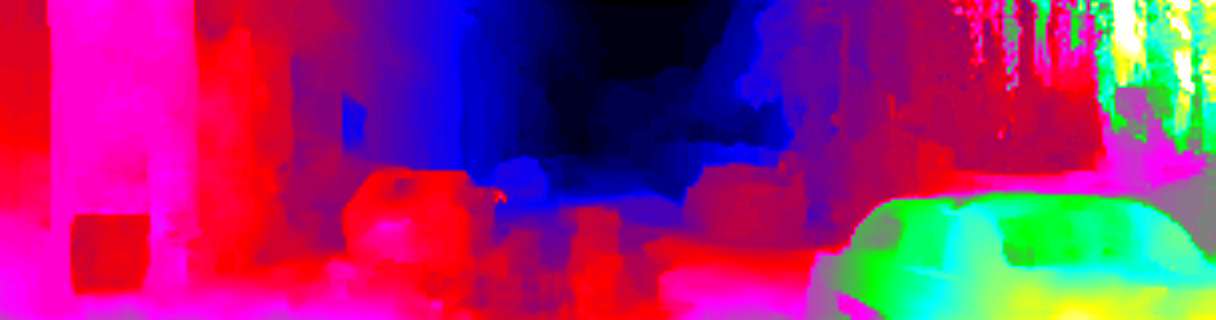} &
		\includegraphics[align=c,width=0.30\textwidth]{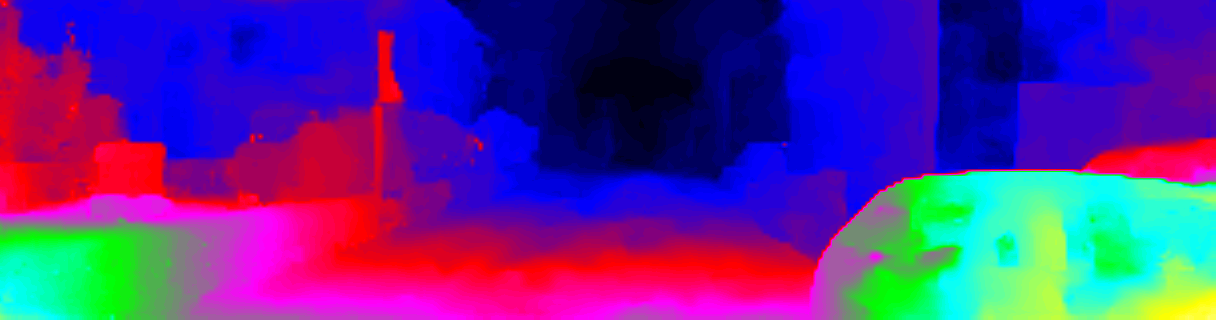}\\
		
		\vspace{1pt}
		&$0^{th}$ frame & $150^{th}$ frame & $300^{th}$ frame\\
		
	\end{tabular}
	\caption{Disparity maps predicted by \netname{} on a \kitti{} sequence  \cite{KITTI_RAW}. Left images (a), no adaptation (b), online adaptation of the \emph{whole} network (c), online adaptation by \algoname{} (d). Green pixel values indicate larger disparities (\ie, closer objects).}
	\label{fig:teaser}
\end{figure*}

We propose to address the domain shift issue by casting \emph{adaptation} as a \emph{continuous learning} process whereby a stereo network can evolve \emph{online}  based on the images gathered by the camera during its real deployment. We believe that the ability to continually adapt itself in real-time is key to any deep learning machinery intended to work in real scenarios. We achieve continuous online adaptation by: deploying one of the unsupervised losses proposed in literature (\ie, \cite{garg2016unsupervised,godard2017unsupervised,Tonioni_2017_ICCV,zhang2018activestereonet}); computing error signals on the current frames; updating the whole network by back-propagation (from now on shortened as \emph{back-prop}); and moving to the next pair of input frames. 
However, such adaptation reduces inference speed greatly. Therefore, to keep a high enough frame rate we propose a novel \extendednetname{} (\netname{}) architecture designed to be lightweight, fast and modular. This architecture exhibits accuracy comparable to DispNetC \cite{mayer2016large} using one-tenth parameters, runs at around  $40$ FPS for disparity inference and performs an online adaptation of the whole network at around $15$ FPS. 
Moreover, to achieve  an even higher frame rate during adaptation, at the cost of a slight loss in accuracy, we develop a \extendedalgoname{} (\algoname{}) algorithm that leverages the modular architecture of \netname{} in order to train sub-portions of the whole network independently. Using \netname{} together with \algoname{} we can adapt our network to unseen environments without supervision at approximately $25$ FPS. 

\autoref{fig:teaser} shows the disparity maps predicted by \netname{} on three successive frames of a video sequence from the \kitti{} dataset \cite{KITTI_RAW}: without undergoing any adaptation - row (b); by adapting online the \emph{whole} network - row (c); and by our computationally efficient \algoname{} approach - row (d). Rows (c) and (d) show how online adaptation can improve the quality of the predicted disparity maps significantly in as few as 150 frames (\ie, a latency of about 10 seconds for complete online adaptation and 6 seconds for \algoname{}).
Extensive experimental results support our three main novel contributions:

\begin{itemize}
\item We cast adaptation as an online task instead of a phase prior to deployment, as previously proposed in \cite{Tonioni_2017_ICCV,pang2018zoom}. We prove that, despite a transition phase, performance of popular networks \cite{mayer2016large} with adaptation are comparable to extensive offline fine-tuning.
    
\item We propose an extremely fast, yet accurate network for stereo matching, \netname{}. Compared to the fastest model in literature \cite{khamis2018stereonet}, \netname{} ranks higher on the online KITTI leader-board \cite{KITTI_2015} and runs faster on the low power NVIDIA Jetson TX2. Moreover, compared to DispNetC, \netname{} adapts better to unseen environments. 
    
\item We propose \algoname{}, a novel training paradigm suited to \netname{} that trades accuracy for speed and allows for significantly faster online adaptation (\ie, 25FPS). Despite this, given sufficiently long sequences, we can achieve comparable accuracy while keeping the speed advantage.
\end{itemize}

To the best of our knowledge, the synergy between \netname{} and \algoname{} realizes the first-ever real-time, self-adapting, deep stereo system.

\section{Related work}
\label{sec:related}


\begin{figure*}[t]
	\centering
	\includegraphics[width=1\linewidth]{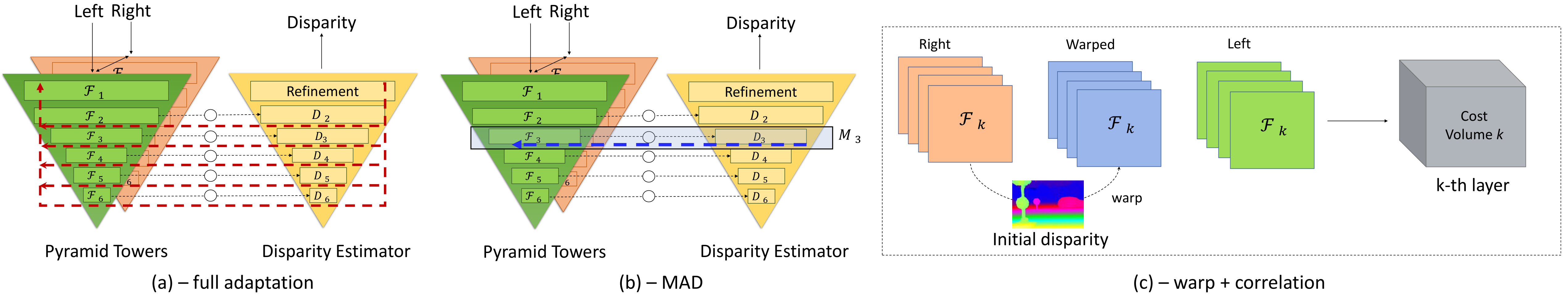} \\
	\caption{Sketch of \netname{} architecture (a), each circle between an $F_k$ and $D_k$ represents a warp and correlation layer (c). Each pair ($\mathcal{F}_i$,$\mathcal{D}_i$) composes a module $\mathcal{M}_i$, adaptable independently by \algoname{},blue arrow in (b), faster than full back-prop, red arrow in (a).}
	\label{fig:architecture}
\end{figure*}

\textbf{Machine learning for stereo.} Early attempts to leverage machine learning for stereo matching concerned estimating confidence measures \cite{Poggi_2017_ICCV}, by random forest classifiers \cite{Hausler_2013_CVPR,Spyropoulos_2014_CVPR,Park_2015_CVPR,Poggi_2016_3DV} and -- later -- by CNNs \cite{Poggi_2016_BMVC,Seki_2016_BMVC,TOSI_2018_ECCV}, typically plugged into conventional pipelines to improve accuracy. 
CNN based matching cost functions \cite{zbontar2016stereo,Chen_2015_ICCV,luo2016efficient} achieved state-of-the-art on both KITTI and Middlebury v3 by replacing conventional cost functions \cite{SCHARSTEIN_COST} within the SGM pipeline  \cite{hirschmuller2005accurate}. Eventually, Shaked and Wolf \cite{Shaked_2017_CVPR} proposed to rely on deep learning for both  matching cost computation and disparity selection, while Gidaris and Komodakis \cite{Gidaris_2017_CVPR} for refinement.
Mayer \etal{} \cite{mayer2016large} proposed the first end-to-end stereo architecture. Although not achieving state-of-the-art accuracy, this seminal work turned out quite disruptive compared to the traditional stereo paradigm outlined in \cite{scharstein2002taxonomy}, highlighting the potential for a totally new approach. Thereby, \cite{mayer2016large} ignited the spread of end-to-end stereo architectures \cite{Kendall_2017_ICCV,Pang_2017_ICCV_Workshops,liang2018learning,chang2018pyramid,jie2018left,guo2019group} that quickly outmatched any other technique on the \kitti{} benchmarks by leveraging on a peculiar training protocol. In particular, the deep network is initially trained on a large amount of synthetic data with groundtruth labels \cite{mayer2016large} and then fine-tuned on the target domain (\eg, \kitti{}) based on stereo pairs with groundtruth. All these contributions focused on accuracy, only recently Khamis \etal{} \cite{khamis2018stereonet} proposed a deep stereo model with a high enough frame rate to qualify for online usage at the cost of sacrificing accuracy. We will show how in our \netname{} this tradeoff is more favourable. 
Unfortunately, all those models are particularly data dependent and their performance dramatically decay when running in environments different from those observed at training time, as shown in \cite{Tonioni_2017_ICCV}. Batsos \etal{} \cite{batsos2018cbmv} soften this effect by combining traditional functions and confidence measures \cite{Hu_2012_PAMI,Poggi_2017_ICCV} within a random forest framework, proving better generalization compared to CNN-based method \cite{zbontar2016stereo}. Finally, guiding end-to-end CNNs with external depth measurements (\eg Lidar) allows for reducing the domain-shift effect, as reported in \cite{POGGI_2019_CVPR}.

\textbf{Image reconstruction for unsupervised learning}.
A recent trend to train depth estimation networks in an unsupervised manner relies on image reconstruction losses. In particular, for monocular depth estimation this is achieved by warping different views, coming from stereo pairs or image sequences, and minimizing the reconstruction error \cite{garg2016unsupervised,zhou2017unsupervised,godard2017unsupervised,zhang2018activestereonet,pydnet18,3net18,Tosi_2019_CVPR}. This principle has also been used for optical flow \cite{Meister:2018:UUL} and stereo \cite{zhou2017unsupervisedStereo}. For the latter task,  alternative unsupervised learning approaches consist in deploying traditional stereo algorithms and confidences \cite{Tonioni_2017_ICCV} or combining by iterative optimization the predictions obtained at multiple resolutions \cite{pang2018zoom}. However, we point out that both works have addressed offline training only, while we propose to solve the very same problem casting it as an online (thus fast) adaptation to unseen environments. 

\section{Online Domain Adaptation}
\label{sec:online}

Modern machine learning models reduce their accuracy when tested on data significantly different from the training set, an issue commonly referred to as \emph{domain shift}. Despite all the research work to soften this issue, the most effective practice still relies on additional offline training on samples from the target environments.
The domain shift curse is inherently present in deep stereo networks since most training iterations are performed on synthetic images quite different from real ones. Then, adaptation can be effectively achieved by fine-tuning the model offline on samples from the target domain by relying on expensive annotations or  unsupervised loss functions \cite{garg2016unsupervised,godard2017unsupervised,Tonioni_2017_ICCV,zhang2018activestereonet}. 

In this paper we move one step further arguing that adaptation can be effectively performed online as soon as new frames are available, thereby obtaining a deep stereo system capable of adapting itself dynamically. For our online adaptation strategy we do not rely on the availability of ground-truth annotations and, instead, use one of the proposed unsupervised losses. To adapt the model we perform on-the-fly a single train iteration (forward and backward pass) for each incoming stereo pair. Therefore, our model is always in training mode and continuously fine-tuning to the sensed environment.

\subsection{MADNet - \extendednetname{}}
\label{sec:network}

One of the main limitations that have prevented exploration of online adaptation is the computational cost of performing a full train iteration for each incoming frame. Indeed, we will show experimentally how it roughly corresponds to a reduction of the inference rate of the system to roughly one third, a price far too high to be paid with most modern architectures. To address this issue, we have developed \extendednetname{} (\netname{}), a novel lightweight model for depth estimation inspired by fast, yet accurate, architectures proposed for optical flow \cite{Ranjan_2017_CVPR,sun2018pwc}. 

We deploy a pyramidal strategy for dense disparity regression for two key purposes: i) maximizing speed and ii) obtaining a modular architecture as depicted in \autoref{fig:architecture}. 
Two pyramidal towers extract features from the left and right frames through a cascade of independent modules sharing the same weights. Each module consists of convolutional blocks aimed at reducing the input resolution by two $3 \times 3$ convolutional layers, respectively with stride 2 and 1, followed by Leaky ReLU non-linearities. According to \autoref{fig:architecture}, we count 6 blocks providing  us with feature $\mathcal{F}$ from half resolution to ${1}/{64}$, namely $\mathcal{F}_1$ to $\mathcal{F}_6$, respectively. These blocks extract 16, 32, 64, 96, 128 and 192 features.

At the lowest resolution (\ie, $\mathcal{F}_6$), we forward features from left and right images into a correlation layer \cite{mayer2016large} to get the raw matching costs. Then, we deploy a disparity decoder $\mathcal{D}_6$ consisting of 5 additional $3 \times 3$ convolutional layers,  with 128, 128, 96, 64, and 1 output channels. Again, each layer is followed by Leaky ReLU, except the last one, which provides the disparity map at the lowest resolution. 

Then, $D_6$ is up-sampled to level $5$ by bilinear interpolation and used both for warping right features towards left ones before computing correlations and as input to $\mathcal{D}_{5}$. Thanks to our design, from $\mathcal{D}_5$ onward, the aim of the disparity decoders $\mathcal{D}_{k}$ is to refine and correct the up-scaled disparities coming from the lower resolution. In our design, the correlation scores computed between the original left and right features aligned according to the lower resolution disparity prediction guide the network in the refinement process. We compute all correlations inside our network along a [-2,2] range of possible shifts.

This process is repeated up to quarter resolution (\ie, $\mathcal{D}_2$), where we add a further refinement module  consisting  of $3 \times 3$ dilated convolutions \cite{sun2018pwc}, with, respectively 128, 128, 128, 96, 64, 32, 1 output channels and 1, 2, 4, 8, 16, 1, 1 dilation factors, before bilinearly upsampling to full resolution. Additional details on the \netname{}  architecture are provided in the supplementary material.

\netname{} has a smaller memory footprint and delivers disparity maps much more rapidly than other more complex networks such as \cite{Kendall_2017_ICCV,chang2018pyramid,liang2018learning} with a small loss in accuracy.
Concerning efficiency, working at decimated resolutions allows for computing correlations on a small horizontal window \cite{sun2018pwc}, while warping features and forwarding disparity predictions across the different resolutions enables to maintain a small search range and look for residual displacements only. 
With a 1080Ti GPU, \netname{} runs at about 40 FPS at \kitti{} resolution and can perform online adaptation with full back-prop at 15 FPS. 

\subsection{MAD - \extendedalgoname{}}
\label{sec:algoDescription}

As we will show, \netname{} is remarkably accurate with full online adaptation at 15 FPS. However, for some applications, it might be desirable to achieve a higher frame rate without losing the adaptation ability. Most of the time needed to perform online adaptation is spent executing back-prop and weights update across all the network layers. A naive way to speed up the process will be to \textit{freeze} the initial part of the network and fine tune only a subset of $k$ final layers, thus realizing a shorter back-prop that would yield a higher frame rate. However, there is no guarantee that these last $k$ layers are indeed those that would benefit most from online fine-tuning. For example, the initial layers of the network should be probably adapted alike, as they directly interact with the images from a new, \textit{unseen}, domain. In \autoref{ssec:strategy} we will provide experimental results to show that training only the final layers is not enough for handling the drastic domain changes that typically occur in practical applications. 

Following the key intuition that to keep up with fast inference we should pursue a partial, though effective, online adaptation, we developed  \extendedalgoname{} (\algoname{}) an online adaptation algorithm tailored to \netname{}, though possibly extendable to any multi-scale inference network. Our method takes a network $\mathcal{N}$ and subdivides it into $p$ non-overlapping portions, each referred to as module $\mathcal{M}_i$, $i \in [1,p]$, such that  $\mathcal{N}=[\mathcal{M}_1,\mathcal{M}_2,..\mathcal{M}_p]$. Each $\mathcal{M}_i$ ends with a final layer able to output a disparity estimation $y_i$. 
Thanks to its design, decomposing our network is straightforward by grouping layers working at the same resolution $i$ from both $\mathcal{F}_i$ and $\mathcal{D}_i$ into a single module $\mathcal{M}_i$, \eg{}, $\mathcal{M}_3 = (\mathcal{F}_3,\mathcal{D}_3$). At each training iteration, thus, we can optimize one of the modules independently from the others by using the prediction $y_i$ to compute a loss function and then executing the shorter back-prop only across the layers of $\mathcal{M}_i$. For instance to optimize $\mathcal{M}_3$ we would use $y_3$ to compute a loss function and back-prop only through $\mathcal{D}_3$ and $\mathcal{F}_3$ following the blue path in \autoref{fig:architecture} (b). Conversely, full back-prop would follow the much longer red path in \autoref{fig:architecture} (a). 
This paradigm allows for 

\begin{itemize}
    \item Interleaved optimization of different $\mathcal{M}_i$, thereby approximating full back-prop over time while gaining considerable speed-up.
    \item Fast adaptation of single modules, which instantly provides benefits to the overall accuracy of the whole network thanks to its cascade architecture.
\end{itemize}

At deployment time, for each incoming stereo pair, we run a forward pass to obtain all estimates $[y_1,\dots,y_p]$ at each resolution, then we choose a portion $\theta \in [1,\dots,p]$ of the network to train according to some heuristic and finally update $\mathcal{M}_\theta$ according to a loss computed on $y_\theta$. We consider a valid heuristic any function that outputs a probability distribution among the $p$ modules of $N$ from which we could perform sampling.

\begin{figure}[t]
	\centering
	\includegraphics[width=0.50\textwidth]{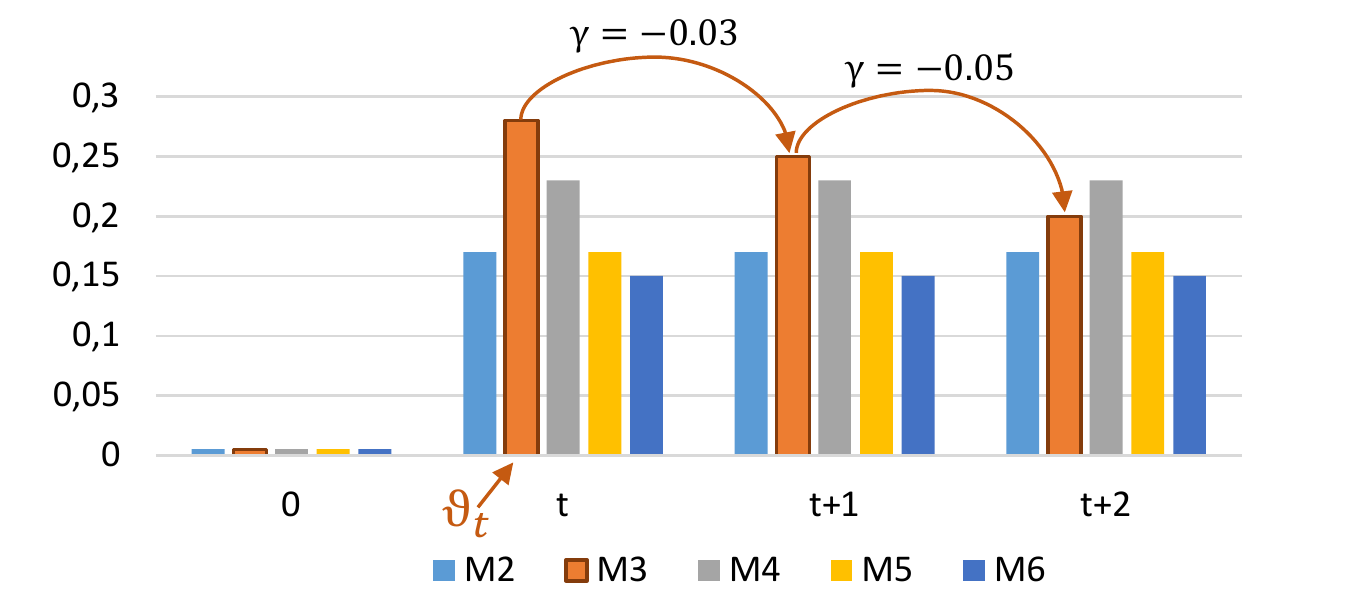}
	\caption{Example of reward/punishment mechanism. $X$ axis shows time while $Y$ histogram values. At time $t$, the most probable module selected for adaptation is $\mathcal{M}_3$. After two steps ($t+2$), its probability gets demoted in favour of $\mathcal{M}_4$.}
	\label{fig:histogram}
\end{figure}

\subsection{Reward/punishment selection}
\label{sec:selectionDescription}

Among different functions, we obtained good results using a reward/punishment mechanism. 
We start by creating a histogram $\mathcal{H}$ with $p$ bins (\ie, one per module) all initialized to 0. For each stereo pair we perform a forward pass to get the disparity predictions $y_i$ and measure the performance of the model by computing a loss $\mathcal{L}_t$ using the full resolution disparity $y$ and  the input frames $x$ (\eg, reprojection error between left and warped right frames as in \cite{godard2017unsupervised}).
Then, we sample the portion to train $\theta \in [1,\dots,p]$ from a probability distribution obtained applying the softmax function to the value of the bins in $(\mathcal{H})$: 

\begin{equation}
\theta_t \sim softmax(\mathcal{H}).
\end{equation}

We can compute one optimization step for layers of $\mathcal{M}_{\theta_t}$ with respect to the loss $\mathcal{L}^{\theta_t}_t$ computed on the lower scale prediction $y_{\theta_t}$. We have now partially adapted the network to the current environment. At the following iteration, we update $\mathcal{H}$ before choosing the new $\theta_t$, increasing the probability of being sampled for the $\mathcal{M}_{\theta_{t-1}}$ that have proven effective. To do so we compute a noisy expected value for $\mathcal{L}_t$ by linear extrapolation of the losses at the previous two-time steps 
\begin{equation}
\mathcal{L}_{exp} = 2 \cdot \mathcal{L}_{t-1}-\mathcal{L}_{t-2},
\end{equation}
and quantify the effectiveness of the last module optimized as 
\begin{equation}
\gamma = \mathcal{L}_{exp}-\mathcal{L}_t.
\end{equation}

Finally, we can change the value of $\mathcal{H}[\theta_t]$ according to $\gamma$, \ie{} effective adaptation will have $\mathcal{L}_{exp} > \mathcal{L}_t$, thus $\gamma > 0$.  
We found out that adding a temporal decay to $\mathcal{H}$ increases the stability of the system, leading to the following update rule

\begin{equation}
\begin{split}
\mathcal{H}&=0.99\cdot\mathcal{H} \\
\mathcal{H}[\theta_{t-1}]&=\mathcal{H}[\theta_{t-1}]+0.01\cdot\gamma
\end{split}
\end{equation}
Additional pseudo code to detail this heuristic is available in the supplementary material.

\autoref{fig:histogram} shows an example of histogram $\mathcal{H}$ at generic time frames $t$ and $t+2$, highlighting the transition from $\mathcal{M}_3$ to $\mathcal{M}_4$ as most probable module thanks to the aforementioned mechanism. 

\section{Experimental results}
\label{sec:results}

\subsection{Evaluation protocol and implementation}
\label{ssec:protocol}

To properly address practical deployment scenarios in which there are no ground-truth data available for fine-tuning in the actual testing environments, we train our stereo network using  synthetic data only \cite{mayer2016large}. More details regarding the training process can be found in the supplementary material. 

To test the online adaptation we use those weights as a common initialization and carry out an extensive evaluation on the large and heterogeneous \kitti{} raw dataset \cite{KITTI_RAW} with depth labels \cite{Uhrig2017THREEDV} converted into disparities by knowing the camera parameters. Overall, we assess the effectiveness of our proposal on 43k images. Specifically, according to the \kitti{} classification, we evaluate our framework in four heterogeneous environments, namely \emph{Road}, \emph{Residential}, \emph{Campus} and \emph{City}, obtained by concatenation of the available video sequences and resulting in 5674, 28067, 1149 and 8027 frames respectively. Although these sequences are all concerned with driving scenarios, each has peculiar traits that would lead deep stereo model to gross errors without  suitable fine-tuning in the target domain. For example, \emph{City} and \emph{Residential} often depict road surrounded by buildings, while \emph{Road}  concerns mostly highways and country roads, where the most common objects are cars and vegetation. 

By processing stereo pairs within sequences, we can measure how well the network adapts, by either full back-prop or \algoname{}, to the target domain compared to a model trained  offline.
For all experiments, we analyze both average End Point Error (EPE) and the percentage of pixels with disparity error larger than 3 (D1-all). Due to the image format being different for each sequence, we extract a central crop of size $320\times1216$ from each frame, which suits to the downsampling factor of our architecture and allows for validating almost all pixels with available ground-truth disparities. 

Finally, we highlight that for both full back-prop and \algoname{},  we compute the error rate on each frame \emph{before} applying the model adaptation step. That is, we measure performances achieved by the current model on the stereo frame at time $t$ and \emph{then} adapt it according to the current prediction. Therefore, the model update carried out at time $t$ will affect the prediction only from frame $t+1$ and so on. 
As unsupervised loss for online adaptation, we rely on the photometric consistency between the left frame and the right one reprojected according to the predicted disparity. Following \cite{godard2017unsupervised}, to compute the reprojection error between the two images we combine the Structural Similarity Measure (SSIM) and the L1 distance, weighted by 0.85 and 0.15, respectively. We selected this unsupervised loss function as it is the fastest to compute among those proposed in literature \cite{Tonioni_2017_ICCV,pang2018zoom,zhou2017unsupervisedStereo} and does not require any additional information besides a pair of stereo images. Further details are available in the supplementary material.

\subsection{MADNet performance}
\label{sec:expr_network}

Before assessing the performance obtainable through online adaptation, we test the effectiveness of \netname{} by following the canonic two-phase training using synthetic \cite{mayer2016large} and real data. Thus, after training on synthetic data, we perform fine-tuning on the training sets of \kitti{} 2012 and \kitti{} 2015 and submit to the \kitti{} 2015 online benchmark. Additional details on the fine-tuning protocol are provided in the supplementary material. On \autoref{tab:submission} we report our result compared to other (published) fast inference architectures on the leaderboard (runtime measured on NVIDIA 1080Ti) as well as with a slower and more accurate one, GWCNet \cite{guo2019group}. At the time of writing, our method ranks 90$^{th}$. Despite the mid-rank achieved in terms of absolute accuracy,  \netname{} compares favorably to StereoNet \cite{khamis2018stereonet} ranked 92$^{nd}$, the only other high frame rate proposal on the KITTI leaderboard. Moreover, we get close to the performance of the original DispNetC \cite{mayer2016large} while using $\frac{1}{10}$ of the parameters and running  more than twice faster. 

\begin{table}[]
    \centering
    \scalebox{0.88}{
    \renewcommand{\tabcolsep}{2pt} 
    \begin{tabular}{|c|c|c|c|c|}
        \hline
         & GWCNet \cite{guo2019group} & DispNetC \cite{mayer2016large} & StereoNet \cite{khamis2018stereonet} & \netname{} \\
        \hline
        D1-all & 2.11 & 4.34 & 4.83 & 4.66 \\
        Time & 0.32 & 0.06 & 0.02 & 0.02 \\
         \hline
    \end{tabular}
    }
    \caption{Comparison between stereo architectures on the \kitti{} 2015 test set without adaptation. Detailed results available in the KITTI online leader-board.}
    \label{tab:submission}
\end{table}

\subsection{Online adaptation}
\label{sec:exp_adaptation}

\begin{table*}[t]
	\center
	\setlength{\tabcolsep}{7pt}
	\scalebox{0.9}
	{
	\begin{tabular}{|c|c|cc|cc|cc|cc|c|}
		\cline{3-10}
		\multicolumn{2}{c}{} & \multicolumn{2}{|c|}{City} & \multicolumn{2}{c|}{Residential} & \multicolumn{2}{c|}{Campus} & \multicolumn{2}{c|}{Road} \\
		\hline
		Model & Adapt. & D1-all(\%) & EPE & D1-all(\%) & EPE & D1-all(\%) & EPE & D1-all(\%) & EPE & FPS\\ 
		\hline
		DispNetC & No & 8.31 & 1.49 & 8.72 & 1.55 & 15.63  & 2.14 & 10.76  & 1.75 & 15.85\\ 
		DispNetC & Full & 4.34  & 1.16  & 3.60 & 1.04 & 8.66  & 1.53 & 3.83 & 1.08 & 5.22 \\
		DispNetC-GT & No & 3.78 & 1.19  & 4.71  & 1.23  & 8.42 & 1.62 & 3.25 & 1.07  & 15.85 \\
		\hline
		\netname{} & No & 37.42 & 9.96 & 37.41 & 11.34 & 51.98 & 11.94 & 47.45 & 15.71 & 39.48\\
		\netname{} & Full & 2.63 & 1.03 & 2.44 & 0.96 & 8.91 & 1.76 & 2.33 & 1.03 & 14.26\\ 
		\netname{} & \algoname{} & 5.82 & 1.51 & 3.96 & 1.31 & 23.40 & 4.89 & 7.02 & 2.03 & 25.43\\ 
		\netname{}-GT & No & 2.21 & 0.80 & 2.80 & 0.91 & 6.77 & 1.32 & 1.75 & 0.83 & 39.48\\
		\hline
	\end{tabular}
	}
	\caption{Performance on the \emph{City}, \emph{Residential}, \emph{Campus} and \emph{Road} sequences from KITTI \cite{KITTI_RAW}. Experiments with DispNetC \cite{mayer2016large} (top) and \netname{} (bottom) with and without online adaptations. \emph{-GT} variants are fine-tuned on KITTI training set groundtruth.}
	\label{tab:sequences}
\end{table*}

We will now show how online adaptation is an effective paradigm, comparable, or better, to offline fine-tuning. 
\autoref{tab:sequences} reports extensive experiments on the four different KITTI environments. We report results achieved by i) DispNetC \cite{mayer2016large} implemented in our framework and trained, from top to bottom, on synthetic data following authors' guidelines, using online adaptation or fine-tuned on groundtruth and ii) \netname{} trained with the same modalities and, also, using \algoname{}. These experiments, together to \autoref{sec:expr_network}, support the three-fold claim of this work.

\textbf{DispNetC: Full adaptation.} On top of \autoref{tab:sequences}, focusing on the D1-all metric, we can notice how running full back-prop online to adapt DispNetC \cite{mayer2016large} decimates the number of outliers on all scenarios compared to the model trained on the synthetic dataset only. In particular, this approach can consistently halve D1-all on \emph{Campus}, \emph{Residential} and \emph{City} and nearly reduce it to one third on \emph{Road}. Alike, the average EPE drops significantly across the four considered environments, with improvement as high as a nearly $40\%$ relative improvement on the Road sequences. These massive gains in accuracy, though, come at the price of slowing the network down significantly to about one-third of the original inference rate, \ie{} from nearly 16 to 5.22 FPS. 
As mentioned above, the Table also reports the performance of the models fine-tuned offline on the 400 stereo pairs with groundtruth disparities from the \kitti{} 2012 and 2015 training dataset \cite{KITTI_2015,KITTI_2012}. It is worth pointing out how online adaptation by full back-prop turns out competitive to fine-tuning offline by groundtruth, and even more accurate in the Residential environment. This fact may hint at training usupervisedly by a more considerable amount of data possibly delivering better models than supervision by fewer data.

\textbf{\netname{}: Full adaptation.} On bottom of \autoref{tab:sequences} we repeat the aforementioned experiments for \netname{}. Due to the much higher errors yielded by the model trained on synthetic data only, full online adaptation turns out even more beneficial with \netname{}, leading to a model which is more accurate than DispNetC with Full adaptation in all sequences but \textit{Campus} and can run nearly three times faster (\ie{} at 14.26 FPS compared to the 5.22 FPS of DispNetC-Full). These results also highlight the inherent effectiveness of the proposed \netname{}. Indeed, as vouched by the rows dealing with  \netname{}-GT and DispNetC-GT, using for both our implementations and training them following the same standard procedure in the field (\ie, pretraining on synthetic data and fine-tuning on KITTI training sets), \netname{} yields better accuracy than DispNetC while running about 2.5 times faster. 

\textbf{\netname{}: \algoname{}.} Once proved that online adaptation is feasible and beneficial, we show that \netname{} employing \algoname{} for adaptation (marked as \algoname{} in column \textit{Adapt.}) allows for effective and efficient adaptation. 
Since the proposed heuristic has a non-deterministic sampling step, we have run the tests regarding \algoname{} five times each and reported here the average performance. We refer the reader to \autoref{ssec:strategy} for analysis on the standard deviation across different runs.
Indeed, \algoname{} provides a significant improvement in all the performance figures reported in the table compared to the corresponding models trained by synthetic data only. 
Using \algoname{},  \netname{}  can be adapted paying a relatively small computational overhead which results in a remarkably fast inference rate of about 25 FPS. 
Overall, these results highlight how, whenever one has no access to training data from the target domain beforehand, online adaptation is feasible and worth. Moreover, if speed is a concern \netname{} combined with \algoname{} provides a favourable trade-off between accuracy and efficiency.       

\begin{figure*}
	\centering
	\includegraphics[width=0.95\textwidth]{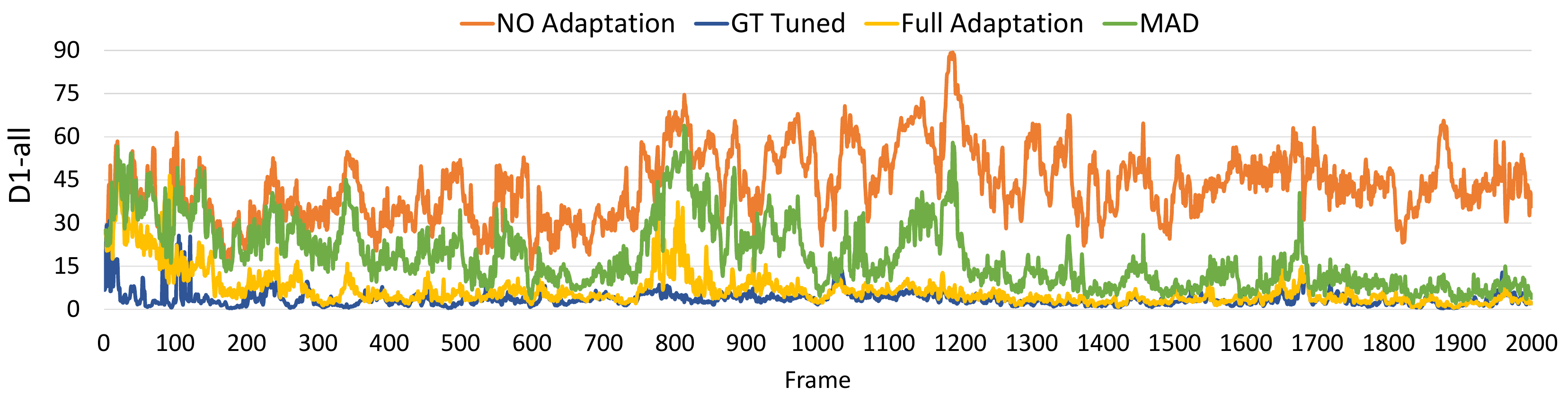}
	\caption{\netname{}: error across frames in the \emph{2011\_09\_30\_drive\_0028\_sync} sequence (\kitti{} dataset, \emph{Residential} environment).
		\label{fig:errorSequence}
	}
\end{figure*}

\textbf{Short-term Adaptation.}
\autoref{tab:sequences} also shows how all adapted models perform significantly worse on \textit{Campus} compared the other sequences. We ascribe this mainly to  \textit{Campus} featuring fewer frames (1149) compared the other sequences (5674, 28067, 8027), which implies a correspondingly lower number of adaptation steps executed online. Indeed, a key trait of online adaptation is the capability to improve performance as more and more frames are sensed from the environment. This favourable behaviour, not captured by the average error metrics reported in \autoref{tab:sequences}, is highlighted in \autoref{fig:errorSequence}, which plots the D1-all error rate over time for \netname{} models in the four modalities. While without adaptation the error keeps being always large, models adapted online clearly improve over time such that, after a certain delay, they become as accurate as the model that could have been obtained by offline fine-tuning had groundtruth disparities been available. In particular, full online adaptation achieves performance comparable to fine-tuning by the groundtruth after 900 frames (\ie, about 1 minute) while for \algoname{} it takes about 1600 frames (\ie, 64 seconds) to reach an almost equivalent performance level while providing a substantially higher inference rate ($\sim25$ vs $\sim15$).

\textbf{Long-term Adaptation.} As \autoref{fig:errorSequence} hints, online adaptation delivers better performance processing a higher number of frames.  In \autoref{tab:overall} we report additional results obtained by concatenating together the four environments without network resets to simulate a stereo camera traveling across different scenarios for $\sim43000$ frames. Firstly,  \autoref{tab:overall} shows how both DispNetC and \netname{} models adapted online by full back-prop yield much smaller average errors than in \autoref{tab:sequences}, as small, indeed, as to outperform the corresponding models fine-tuned offline by groundtruth labels. Hence, performing online adaptation through long enough sequences, even across different environments, can lead to more accurate models than offline fine-tuning on few samples with groundtruth, which further highlights the great potential of our proposed \emph{ continuous learning} formulation. Moreover, when leveraging on \algoname{} for the sake of run-time efficiency, \netname{} attains larger accuracy gains through \emph{continuous learning} than before (\autoref{tab:overall} vs. \autoref{tab:sequences}) shrinking the performance gap between \algoname{} and Full back-prop. We believe that this observation confirms the results plotted in \autoref{fig:errorSequence}: \algoname{} needs more frame to adapt the network to a new environment, but given sequences long enough can successfully approximate full back propagation over time (\ie, 0.20 EPE difference and 1.2 D1-all between the two adaptation modalities on \autoref{tab:overall}) while granting nearly twice FPS. On long term (e.g., beyond 1500 frames on \autoref{fig:errorSequence}) running \algoname{}, full adaptation or offline tuning on groundtruth grants equivalent performance. 
Besides \autoref{fig:teaser}, we report qualitative results in the supplementary material as two video sequences regarding outdoor \cite{KITTI_RAW} and indoor \cite{indoorDataset} environments.

\begin{table}[t]
	\setlength{\tabcolsep}{9pt}
	\center
    \scalebox{0.82}
    {
		\begin{tabular}{c}
			\begin{tabular}{|c|c|cc|c|}
				\hline
				Model & Adapt. & D1-all(\%) & EPE & FPS \\ 
				\hline
				DispNetC & No & 9.09 & 1.58  & 15.85 \\
				DispNetC & Full & 3.45 & 1.04   & 5.22  \\
				DispNetC-GT & No & 4.40 & 1.21 & 15.85 \\
				\hline
				\netname{} & No & 38.84 & 11.65 & 39.48 \\
				\netname{} & Full & 2.17 & 0.91 & 14.26 \\ 
				\netname{} & \algoname{} & 3.37 & 1.11 & 25.43\\ 
				\netname{}-GT & No & 2.67 & 0.89 & 39.48 \\
				\hline
			\end{tabular} \\
	\end{tabular}
	}
	\caption{Results on the full KITTI raw dataset \cite{KITTI_RAW} (\emph{Campus} $\rightarrow$ \emph{City} $\rightarrow$ \emph{Residential}  $\rightarrow $ \emph{Road}).}
	\label{tab:overall}
\end{table}

\subsection{Additional results}
\label{ssec:additional}

\begin{figure}
    \centering
    \begin{tabular}{c}
     \includegraphics[width=1\linewidth]{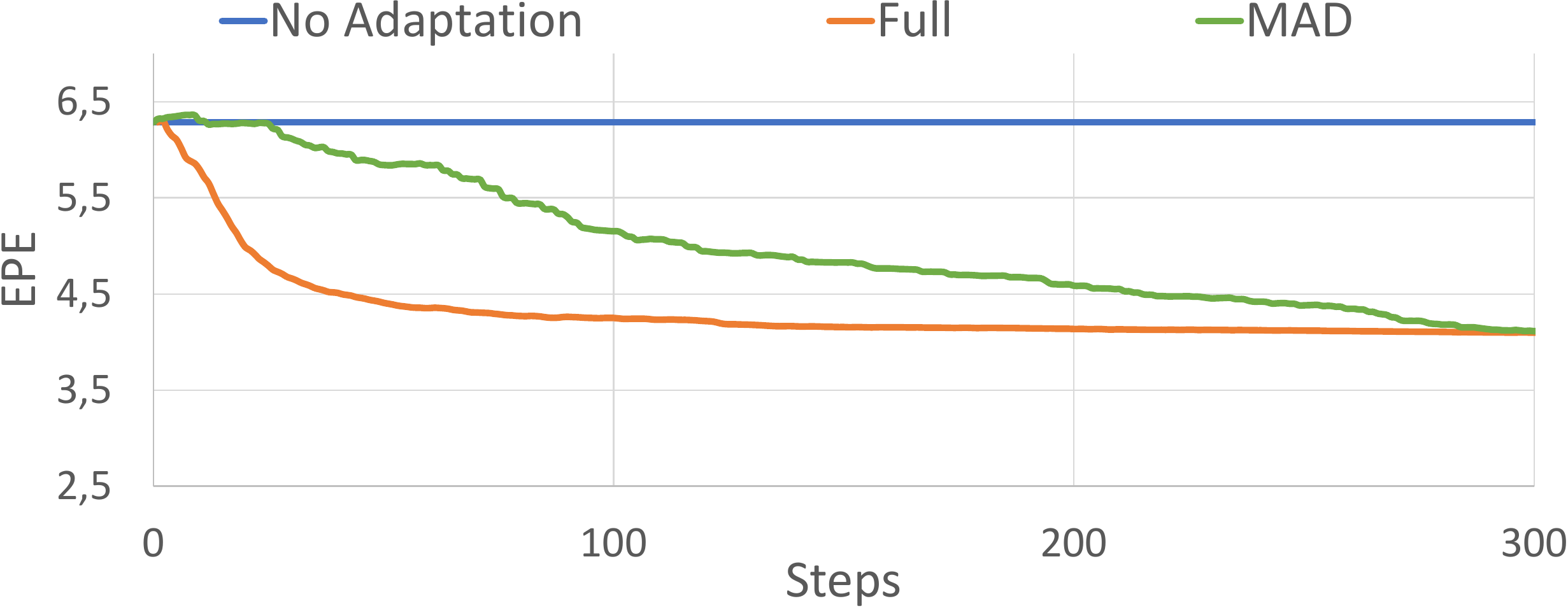} \\    
    \includegraphics[width=1\linewidth]{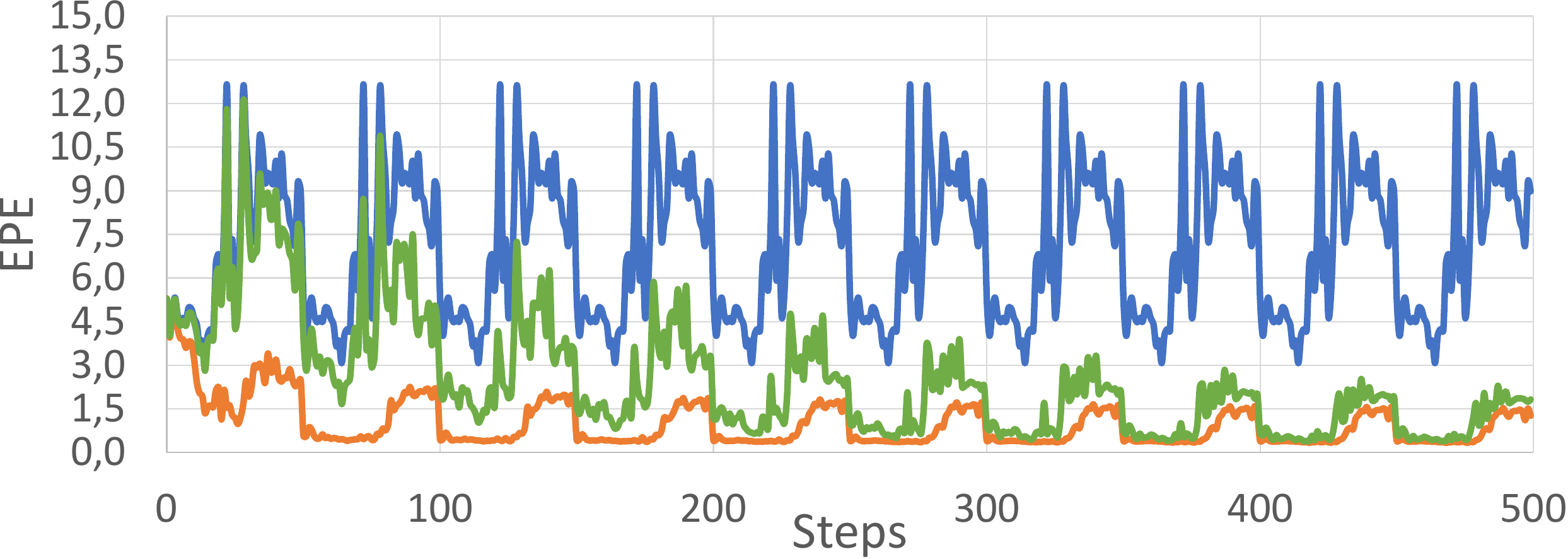} \\
    \end{tabular}
    \caption{End-Point Error (EPE) on Middlebury \emph{Motorcycle} pair (top) and Sintel \emph{Alley-2} sequence (bottom) looped over 10 times.}
    \label{fig:plots}
\end{figure}

Here we show the generality of \algoname{} on environments different from those depicted in the KITTI dataset. To this purpose, we run aimed experiments on the Sintel \cite{Butler:ECCV:2012} and Middlebury \cite{MIDDLEBURY_2014} datasets and plot EPE trends for both \emph{Full} and \emph{Mad} adaptation on \autoref{fig:plots}.
This evaluation allows for measuring the performance on a short sequence concatenated multiple times (\ie, Sintel) or when adapting on the same stereo pair (\ie, Middlebury) over and over.

On Middlebury (top) we perform 300 steps of adaptation on the \textit{Motorcycle} stereo pair.
The plots clearly show how \algoname{} converges to the same accuracy of Full after around 300 steps while maintaining real-time processing (25.6 FPS on image scaled to a quarter of the original resolution).
On Sintel (bottom), we adapt to the \emph{Alley-2} sequence looped over 10 times.
We can notice how the very few, \ie{} 50, frames of the sequence are not enough to achieve good performance with \algoname{}, since it performs the best on long-term adaptation as highlighted before. However, by looping over the same sequence, we can perceive how \algoname{} gets closer to full adaptation, confirming the behavior already experimented on the KITTI environments.

\subsection{Different online adaptation strategies}
\label{ssec:strategy}

\begin{table}[]
    \centering
    \scalebox{0.82}{
    \begin{tabular}{|c|cc|c|}
        \hline
        Adaptation Mode & D1-all(\%) & EPE & FPS  \\
        \hline
        No & 38.84 & 11.65 & 39.48 \\
        \emph{Last layer} & 38.33 & 11.45 & 38.25 \\
        \emph{Refinement} & 31.89 & 6.55 & 29.82\\
        D$_2$+\emph{Refinement} & 18.84 & 2.87 & 25.85 \\
        \algoname{}-SEQ & 3.62 & 1.15 & 25.74\\
        \algoname{}-RAND & 3.56 ($\pm 0.05$) & 1.13 ($\pm 0.01$) & 25.77\\
        \algoname{}-FULL & \textbf{3.37 ($\pm 0.1$)} & \textbf{1.11 ($\pm 0.01$)} & 25.43\\        
        \hline
    \end{tabular}
    }
    \caption{Results on the \kitti{} raw dataset \cite{KITTI_RAW} using \netname{} trained on synthetic data and different fast adaptation strategies}
    \label{tab:adaptation_strategy}
\end{table}

We carried out additional tests on the whole \kitti{} RAW dataset \cite{KITTI_RAW} and compared performance obtainable deploying different fast adaptation strategies for \netname{}. Results are reported on \autoref{tab:adaptation_strategy} together with those concerning a network that does not perform any adaptation. 

First, we compared \algoname{} keeping the weights of the initial portions of the network frozen and training only: the last layer, the \emph{Refinement} module or both D$_2$ and \emph{Refinement} modules. 
Then, since \algoname{} consists in splitting the network into independent portions and choosing which one to train, we compare our full proposal (\textit{\algoname{}-FULL}) to keeping the split and choosing the portion to train either randomly (\textit{\algoname{}-RAND}) or using a round-robin schedule (\textit{\algoname{}-SEQ}). Since \textit{\algoname{}-FULL} and \textit{\algoname{}-RAND} feature non-deterministic sampling steps, we report their average performance obtained across 5 independent runs on the whole dataset with the corresponding standard deviations between brackets.

By comparing the first four entries with the ones featuring \algoname{} we can see how training only the final layers is not enough to successfully perform online adaptation. Even training as many as 13 last layers (\ie, $D_2+Refinement$), at a computational cost comparable with \algoname{}, we are at most able to halve the initial error rate, with performance still far from optimal. The three variants of \algoname{} by training the whole network can successfully reduce the D1-all to $\frac{1}{10}$ of the original. Among the three options, our proposed layer selection heuristic provides the best overall performance even taking into account the slightly higher standard deviation caused by our sampling strategy. Moreover, the computational cost to pay to deploy our heuristic is negligible losing only $0.3$ FPS compared to the other two options.

\subsection{Deployment on embedded platforms}
\label{ssec:mobile}

All the tests reported so far have been executed on a PC equipped with an NVIDIA 1080 Ti GPU. Unfortunately, for many application like robotics or autonomous vehicles, it is unrealistic to rely on such high end and power-hungry hardware. However, one of the key benefits of \netname{} is its lightweight architecture conducive to easy deployment on low-power embedded platforms.
Thus, we evaluated \netname{} on an NVIDIA Jetson TX2 when processing stereo pairs at the full KITTI resolution and compared it to StereoNet \cite{khamis2018stereonet} implemented using the same framework (\ie, the same level of optimization). We measured $0.26s$ for a single forward of \netname{} versus $0.76$-$0.96s$ required by StereoNet, with 1 or 3 refinement modules respectively. 
Thus, for embedded applications \netname{} is an appealing alternative to \cite{khamis2018stereonet} since it is both faster and more accurate.


\section{Conclusions and future work}
\label{sec:conclusion}
The proposed online unsupervised fine-tuning approach can successfully tackle the domain adaptation issue for deep end-to-end disparity regression networks.
We believe this to be key to practical deployment of these potentially ground-breaking deep learning systems in many relevant scenarios. For applications in which inference time is critical,  we have proposed \netname{}, a novel network architecture, and  \algoname{}, a strategy to effectively adapt it online very efficiently. We have shown how \netname{} together with \algoname{} can adapt to new environments by keeping a high prediction frame rate (\ie, 25FPS) and yielding better accuracy than popular alternatives like DispNetC. 
As  main topic for future work, we plan to test and possibly extend \algoname{} to any end-to-end stereo system. We would also like to investigate alternative approaches to select the portion of the network to be updated online at each step. 

\textbf{Acknowledgements.} We gratefully acknowledge the support of NVIDIA Corporation with the donation of a Titan X used for this research.


{\small
\bibliographystyle{ieee}
\bibliography{egbib}
}

\newpage\phantom{blabla}


\section*{Supplementary material}

# Supplementary material for "Real-time self-adaptive deep stereo"

Alessio Tonioni, Fabio Tosi, Matteo Poggi, Stefano Mattoccia, Luigi di Stefano
Department of Computer Science and Engineering (DISI)
University of Bologna, Italy
{alessio.tonioni, fabio.tosi5, m.poggi, stefano.mattoccia, luigi.distefano }@unibo.it

## 1. Detailed *MADNet* structure

We report here a detailed description of the three modules that compose *MADNet*. We start from Fig. 1 depicting details of our pyramidal convolutional feature extractor. *MADNet* will deploy two of them with parameter sharing to extract features independently on the left and right frames (green and red pyramid in Figure 2 (a) in the main paper).

Then, for each one of the 6 resolutions considered in *MADNet*, we build one disparity estimation module as described in Fig. 2. Let $\mathcal{D}_n$ be a disparity estimation module for resolution $n$. The first operation performed is the computation of a cost volume (*i.e.*, correlation layer[4]) between corresponding convolutional features at the same resolution extracted from the left and right image ($\mathcal{F}_n^l$ and $\mathcal{F}_n^r$ respectively). If a lower resolution disparity is available (*e.g.*, $\mathcal{D}_{n+1}$), the features from the right image can be partially aligned to the one on the left before the cost volume computation by using a warping layer [5]. With this strategy we aim to encode in the cost volume useful information for the refinement of the lower resolution disparity $\mathcal{D}_{n+1}$. Then the final input to the module is obtained by concatenating the cost volume obtained and the up-scaled lower res disparity. At the lowest resolution in our network ($\mathcal{D}_6$) we ignore the warping and up-sampling steps (since we do not have a $\mathcal{D}_7$) and directly create a cost volume between $\mathcal{F}_6^l$ and $\mathcal{F}_6^r$. At the highest resolution considered by our model ($\mathcal{D}_2$) we deploy a residual refinement module, depicted in Fig. 3. Here we use atrous convolutions and residual connections to get the final disparity estimation at the same resolution as its input. To recover full resolution, we up-sample the output of the refinement module using bi-linear interpolation.

## 2. Implementation and training details for *MADNet* and *MAD*

We resume implementation details regarding how we have initially trained *MADNet* on synthetic data, how we split the network into independent portions and how we compute the self-supervised loss to train them online.

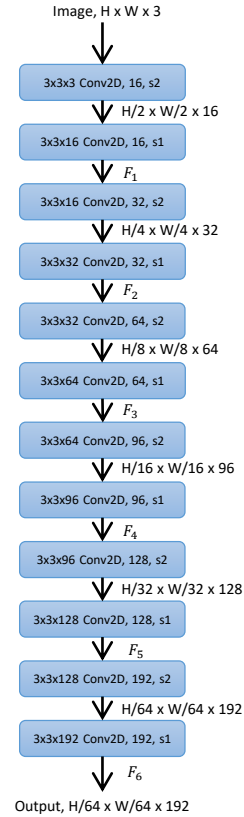

Figure 1. Detailed structure of our convolutional feature extractor. For each convolutional layer, we report the kernel dimensions and the stride as $s$ followed by the stride step.

**Pre-Training:** Regarding the initial training of *MADNet*, we perform 1200000 training iterations on the FlyingThings3D dataset using Adam Optimizer and a learning rate of 0.0001, halved after 400k steps and further every 200k until convergence. As loss function, we compute the sum of per-pixel absolute errors between disparity maps estimated at each resolution and downsampled groundtruth labels. The final loss is a weighted sum of the contributions from the different resolutions. The weights used are respec-

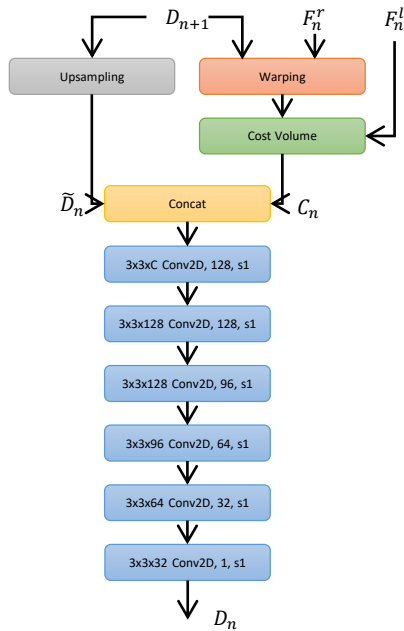 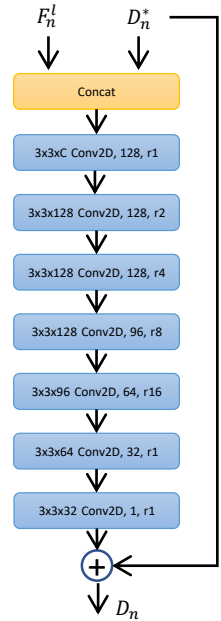

Figure 2. Detailed structure of our stereo estimation network at resolution $n$. We denote with $\mathcal{F}_n$ the convolutional features extracted at resolution $n$, superscript $l$ for those obtained from the left frame and $r$ for those referring to the right one. For the upsampling block, we use standard bilinear upsampling while for the cost volume creation we use the correlation layer introduced in DispNetC [4] with max displacement 2.

Figure 3. Detailed structure of our residual refinement network. The network takes as input an initial disparity estimation $\mathcal{D}_n^*$ and the corresponding left convolutional features at the same resolution $\mathcal{F}_n^l$, then uses atrous convolutions to elaborate them (we report the rate used as $r$ followed by the rate value) and finally outputs a residual pixel-wise correction.

tively 0.005, 0.01, 0.02, 0.08, 0.32 from $\mathcal{D}_2$ to $\mathcal{D}_6$ following [5]. For the additional fine tuning on the KITTI 2012 and 2015 training sets used in section 4.2 of the submitted paper, we performed 500K optimization steps by computing the loss only on the full-resolution disparity map and using 0.001 as weight. All the other hyperparameters are kept as in the synthetic training.

**Adaptation** Concerning *MAD*, we have grouped layers (either from the feature extractor or the disparity estimators) according to the resolution at which they operate, *i.e.* $(\mathcal{F}_i, \mathcal{D}_i)$ composes a module $\mathcal{M}_i$. We made this decision because in our architecture layers at the same resolution are directly connected through skip connections which may allow approximate backpropagation by flowing the gradients only through the connection without traversing the low-resolution layer in between. For example, considering the structure of the disparity estimation module depicted in Fig. 2, we would backprop only through $\mathcal{F}_n^l$ and $\mathcal{F}_N^r$ and not through $\mathcal{D}_{n+1}$. By following this strategy, we obtain 5 independently trainable portions of *MADNet* by grouping layers that produce $[\mathcal{D}_k, \mathcal{F}_k]$ for each one of the resolutions from 6 to 3. For the higher resolution portions (1-2) we collapsed together layers working at the half and quarter resolution (*i.e.*, $\mathcal{F}_1$, $\mathcal{F}_2$, $\mathcal{D}_2$ and the refinement module). Each independent portion of *MADNet* can be trained independently since each one can produce a disparity estimation amenable for loss computation.

We use as loss function for the adaptation the photometric consistency between the left RGB frame of the stereo couple and the right one reprojected according to the predicted disparity. Following [2], we perform the image reprojection using a fully differential bilinear sampler, then compare the two images using a linear combination of SSIM [6] computed on $3 \times 3$ patches and $L_1$ distance. The contribution of the two component are respectively weighted 0.85 and 0.15. We did not use the left-right consistency check proposed in [2] as it would need to elaborate each stereo pair twice, drastically reducing the frame rate of the system without improving the performance considerably. Finally, we run some experiments adding the smoothness term proposed in [2], without getting noticeable improvement. Therefore, we decided to omit it to keep our formulation simpler. For all our tests, every loss function is computed at full resolution by upsampling the small-scale predictions using bilinear sampling. All the code used for our experiments is available [1].

---
[1] https://github.com/CVLAB-Unibo/Real-time-self-adaptive-deep-stereo

2

**Algorithm 1** Online Adaptation with *MAD*

1: **Require:** $\mathcal{N} = [n_1, \ldots, n_p]$
2: $\mathcal{H} = [h_1, \ldots, h_p] \leftarrow 0$
3: **while** *not stop* **do**
4:     $x \leftarrow readFrames()$
5:     $[y, y_1, \ldots, y_p] \leftarrow forward(\mathcal{N}, x)$
6:     $\mathcal{L}_t \leftarrow loss(x, y)$
7:     $\theta \leftarrow sample(argsoftmax(\mathcal{H}))$
8:     $\mathcal{L}_t^\theta \leftarrow loss(x, y_\theta)$
9:     $updateWeights(\mathcal{L}_t^\theta, n_\theta)$
10:     **if** $firstFrame$ **then**
11:         $\mathcal{L}_{t-2} \leftarrow \mathcal{L}_t, \mathcal{L}_{t-1} \leftarrow \mathcal{L}_t, \theta_{t-1} \leftarrow \theta$
12:     **end if**
13:     $\mathcal{L}_{exp} \leftarrow 2 \cdot \mathcal{L}_{t-1} - \mathcal{L}_{t-2}$
14:     $\gamma \leftarrow \mathcal{L}_{exp} - \mathcal{L}_t$
15:     $\mathcal{H} \leftarrow 0.99 \cdot \mathcal{H}$
16:     $\mathcal{H}[\theta_{t-1}] \leftarrow \mathcal{H}[\theta_{t-1}] + 0.01 \cdot \gamma$
17:     $\theta_{t-1} \leftarrow \theta_t, \mathcal{L}_{t-2} \leftarrow \mathcal{L}_{t-1}, \mathcal{L}_{t-1} \leftarrow \mathcal{L}_t$
18: **end while**

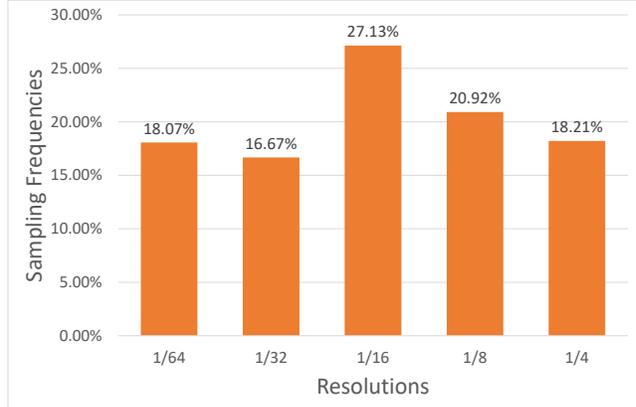

Figure 4. Sampling frequencies for the different independent portions of *MADNet* using *MAD* for fast online adaptation.

## 3. Detailed algorithm for one online adaptation step using *MAD*

Alg. 1 provides detailed pseudocode for online adaptation with *MAD* using our proposed sampling heuristics.

We start by creating a histogram $\mathcal{H}$ with $p$ bins, *i.e.* one per module, all initialized at 0. For each stereo pair we perform a forward pass to get the disparity predictions (line 5) and measure the performance of the model by computing the loss $\mathcal{L}_t$ according to the full resolution disparity $y$ and, potentially, the input frames $x$, *e.g.*, reprojection error between left and right frames [2] (line 6). Then, we pick the portion to train $\theta \in [1, \ldots, p]$ by sampling from the probability distribution obtained as $softmax(\mathcal{H})$ (line 7). Once selected, we compute one optimization step for layers of $\mathcal{M}_\theta$ with respect to the loss $L_t^\theta$ computed on the lower scale prediction $y_\theta$ (line 8-9). We have now partially adapted the network to the current environment. Next, we update $\mathcal{H}$ increasing the probability of being sampled for the $\mathcal{M}_i$ that have proven effective. To do so, we can compute a noisy expected value for $\mathcal{L}_t$ by linear interpolation of the losses at the previous two time step: $L_{exp} = 2 \cdot \mathcal{L}_{t-1} - \mathcal{L}_{t-2}$ (line 13). By comparing it with the measured $\mathcal{L}_t$ we can assess the impact of the network portion sampled at the previous step ($\theta_{t-1}$) as $\gamma = L_{exp} - \mathcal{L}_t$, and then increase or decrease its sampling probability accordingly (*i.e.*, if the adaptation was effective $\mathcal{L}_{exp} > \mathcal{L}_t$, thus $\gamma > 0$). We found out that adding a temporal decay to $\mathcal{H}$ helps increasing the stability of the system, so the final update rule for each step is: $\mathcal{H} = 0.99 \cdot \mathcal{H}, \mathcal{H}[\sigma_{\tau-1}] += 0.01 \cdot \gamma$ (lines 15 and 16).

## 4. *MAD* sampling policy

We are interested in investigating which portions of the network are trained more by *MAD* according to the reward-punishment mechanism we designed. To get some insights we ran *MADNet* performing adaptation with *MAD* on the full KITTI raw dataset 5 times and kept track of the number of steps on which each portion has been sampled for training. In Fig. 4 we report on the y-axis the average sampling frequencies for each portion, whereby on the x-axis we identify each of the portions by the different scale at which it operates. Surprisingly the most sampled portion (*i.e.*, the one that according to our heuristic will grant the greater improvement once adapted) is not the last that produces the final predictions (*i.e.*, $\frac{1}{4}$), but the middle portion of the network (*i.e.*, $\frac{1}{16}$). As pointed out by [3], a good coarse disparity map can be easily up-sampled and refined to an accurate full resolution output. This is further confirmed by our analysis, showing how *MAD* favors the fine-tuning of lower resolution modules(*i.e.*, $\frac{1}{16}$). This behaviour might be closely linked to the architecture of *MADNet* that starts from a low resolution disparity estimation and iteratively refine it. If the low resolution disparity has major mistakes the upper modules are not able to solve them properly, thus training more the lower resolution modules might be preferable.

## 5. Qualitative comparison between fast networks

Fig. 5 provides additional qualitative comparisons between the output of three fast stereo architecture: DispNetC [4], StereoNet [3] and our *MADNet*. We wish to highlight how *MADNet* better maintains thin structures compared to StereoNet.

3

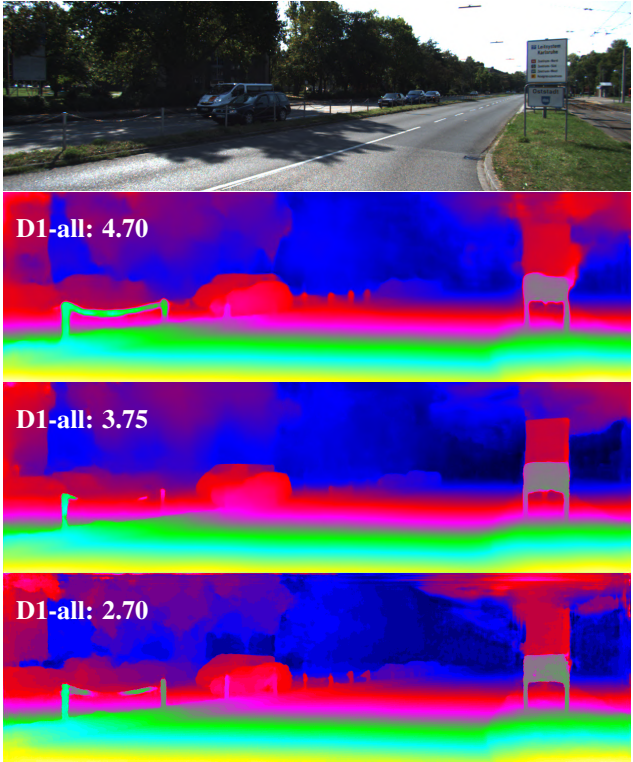

Figure 5. Qualitative comparison between disparity maps from different architectures. From top to bottom, reference image from KITTI 2015 online benchmark and disparity map by DispNetC [4], StereoNet [3] and *MADNet*.

## 6. Qualitative Results on Online Adaptation

As a further supplementary material, we refer the reader to a video showing the effectiveness of our online adaptation formulation in the two different configurations (Full and *MAD*), available at https://www.youtube.com/watch?v=7SjyzDxmCY4. The colormap used encodes with bright color points close to the camera (*i.e.*, high disparity) and with dark colors points far from the camera (*i.e.*, low disparity). We visualize in the upper right corner the predictions performing only inference at 40FPS, in the lower left performing fast adaptation using *MAD* at 25 FPS and in the lower right performing adaptation of the whole network using full back-prop at 15 FPS. For better visualization, we have synchronized the output of the three networks, so the video is not showing real execution times. To give a quantitative measurement of the improvement, we have superimposed over each image in the top left corner the D1-ALL and EPE metrics. For the two adapting networks we also report, between brackets, the gain compared to the same network without adaptation.

For the first half of the video we have select a video sequence from the KITTI raw dataset belonging to the *Residential* environment. The video clearly shows how online adaptation of the whole network by full backprop can solve most of the mistakes in as few as $\sim 150$ frames (*i.e.*, $10s$ of execution time at $15 FPS$). Fast adaptation using *MAD*, instead, requires slightly more frames to achieve comparable improvements (*i.e.*, about $\sim 400$ frames or $16s$ at $25FPS$), but then can still benefit from the higher frame rate. Finally, we can see how going towards the end of the sequence the gap between the adaptation of the whole network by full backprop and *MAD* gets smaller and smaller up to converge to comparable performance in the final $\sim 500$ frames.

The second half of the video concerns performance achieved in an indoor scenario. For this qualitative evaluation we select a sequence from the Wean Hell dataset [1]. We can see how both adaptation strategies can drastically improve the network that does not perform adaptation. As in the outdoor scenario, the full adaptation requires less frame to achieve good performance while *MAD* needs slightly more frames. By the end of the video, both networks produce similar smooth predictions.

Finally, to better highlight some differences between the two adaptation methods we report on Fig. 6 and Fig. 7 some selected frames from the videos. On the left most column we show for each row the index of the frame in the sequence considered. For each example we show the disparity predicted by three different configuration of *MADNet*: without online adaptation, with online adaptation by full back-prop and with our computationally efficient *MAD*.

Fig. 6 shows predictions obtained on a sequence from the KITTI dataset. With as few as $100$ frame *Full Adaptation* is able to resolve most of the mistakes in the predicted disparity, while *MAD* at the same iteration is only able to slightly decrease the magnitude of the mistake. Around the $500^{th}$ iteration (row 2), *MAD* start to improve drastically, with the predicted disparity showing way less mistakes. The same trends is visible in the following rows with *MAD* rapidly closing the performance gap with respect to *Full Adaptation*. By frame 2000 (*i.e.*, after 2000 step of online adaptation) both adaptation techniques converge to similar predictions

Fig. 7 shows predictions obtained on a sequence from the indoor Wean Hall dataset [1]. Even in this scenario both the adaptation mode are able to drastically increase the quality of the predicted disparity maps with relatively few considered steps. In particular by the $500^{th}$ frame the *Full Adaptation* has already adapted the model to the current environment as shown by the absence of macro mistakes in the predicted disparities. *MAD*, instead, needs more iterations, but once again can converge to performance comparable to full adaptation after around $1500$ frames (rows 4 and 5 in Fig. 7). The frames reported in the last two rows show challenging scenes where the re-projection loss that we use for adaptation may fail to produce useful gradients due to the big reflections of the neon lights on the floor that will be

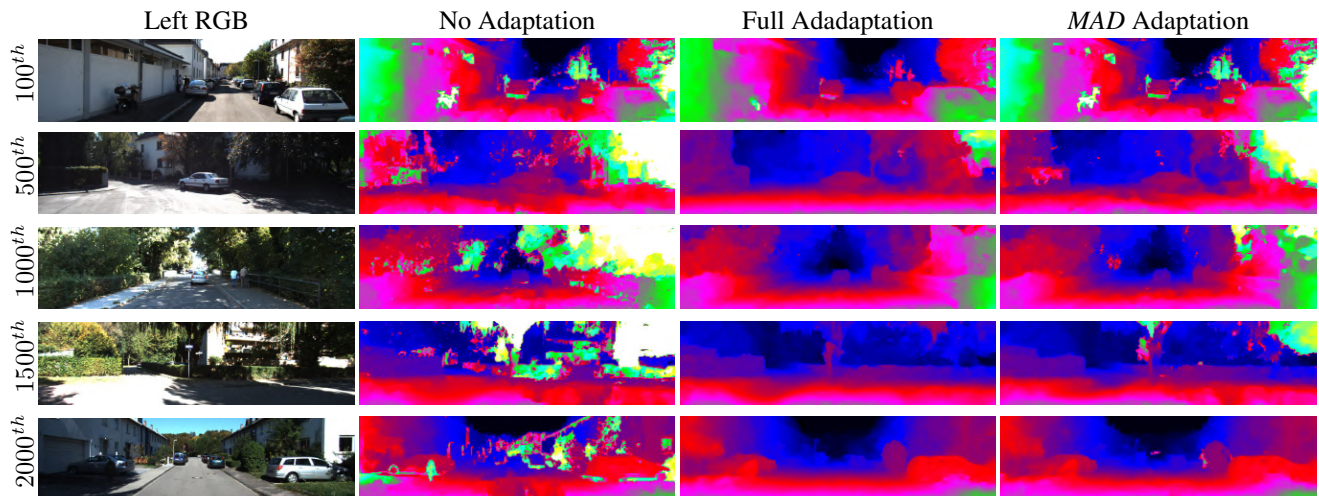

Figure 6. Comparison between predicted disparities obtained by *MADNet* with different adaptation modalities. The leftmost side of the table report the number of elaborated step in the sequence.

viewed differently by the left and right camera. We can see how all the three models fail to produce good prediction in this challenging situation, we plan to address this kind of challenging situation in the future by relying on multiple unsupervised losses.

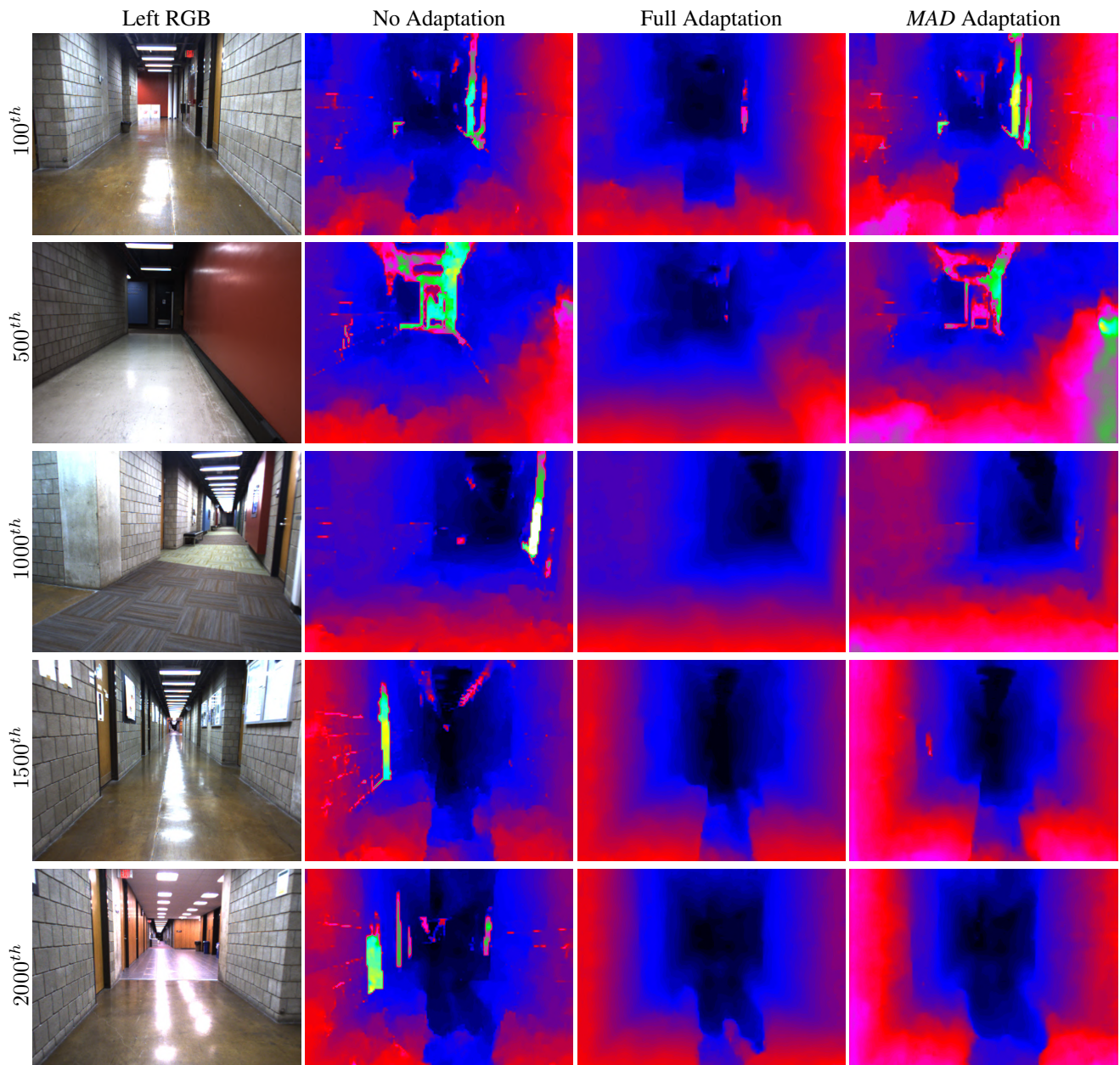

Figure 7. Comparison between predicted disparities obtained by *MADNet* with different adaptation techniques. The leftmost side of the table report the number of elaborated step in the sequence.


\end{document}